\def\year{2020}\relax
\documentclass[letterpaper]{article} %
\usepackage{aaai20}  %
\usepackage{times}  %
\usepackage{helvet} %
\usepackage{courier}  %
\usepackage[hyphens]{url}  %
\usepackage{graphicx} %
\usepackage{balance}
\usepackage{graphicx}  %
\title{Explainable Deep RDFS Reasoner}

\author{
Bassem Makni$^\ddag$,
Ibrahim Abdelaziz$^\ddag$,
James Hendler$^\diamond$
\AND
{\normalsize \rm $^\ddagger$IBM Research} \\
IBM T.J. Watson Research Center \\
$^\dagger$Yorktown Heights, New York 10598, USA \\
\small{\{bassem.makni, ibrahim.abdelaziz1\}@ibm.com} \\
\And
{\normalsize\rm $^\diamond$Rensselaer Polytechnic Institute} \\
Tetherless World Constellation \\
Troy, New York 12180, USA\\
\small{hendler@cs.rpi.edu}
}

\newcommand{\stitle}[1]{\vspace{1ex}\noindent\textbf{#1}}

\usepackage[subpreambles=true]{standalone}
\usepackage[utf8]{inputenc}
\usepackage[english]{babel}
\usepackage{import}
 \usepackage{pgfplots}
\usepackage[acronym]{glossaries}
\newacronym{cuda}{CUDA}{Compute Unified Device Architecture}
\newacronym{lstm}{LSTM}{Long Short Term Memory}
\newacronym{gru}{GRU}{Gated Recurrent Unit}
\newacronym{lubm}{LUBM}{Lehigh University Benchmark}
\newacronym{nel}{NEL}{Named Entity Linking}
\newacronym{ggnn}{GGNN}{Gated Graph Sequence Neural Networks}
\newacronym{gpu}{GPU}{Graphics Processing Unit}
\newacronym{pca}{PCA}{Principal Component Analysis}
\newacronym{sparql}{SPARQL}{SPARQL Protocol and RDF Query Language}
\newacronym{hope}{HOPE}{High-Order Proximity preserved Embedding}
\newacronym{lod}{LOD}{Linked Open Data}
\newacronym{rdf}{RDF}{Resource Description Framework}
\newacronym{rdfs}{RDFS}{RDF Schema}
\newacronym{uri}{URI}{Uniform Resource Identifier}
\newacronym{iri}{IRI}{Internationalized Resource Identifier}
\newacronym{kg}{KG}{Knowledge Graph}
\newacronym{lpg}{LPG}{Labeled Property Graph}
\newacronym{nlp}{NLP}{Natural Language Processing}
\newacronym{rnn}{RNN}{Recurrent Neural Network}
\newacronym{brnn}{BRNN}{Bidirectional Recurrent Neural Network}
\newacronym{idea}{IDEA}{Institute for Data Exploration and Applications}
\newacronym{yasr}{YaSR}{Yet another Semantic Reasoner}
\newacronym{cbow}{CBOW}{Continuous Bag of Words}
\newacronym{lsa}{LSA}{Latent Semantic Analysis}
\newacronym{lda}{LDA}{Latent Dirichlet Allocation}
\newacronym{rtn}{RTN}{Relational Tensor Networks}
\newacronym{rntn}{RNTN}{Recursive Neural Tensor Networks}
\newacronym{dag}{DAG}{Directed Acyclic Graph}
\newacronym{sumo}{SUMO}{Suggested Upper Merged
Ontology}
\newacronym{nell}{NELL}{Never-Ending Language Learning}
\newacronym{kde}{KDE}{Kernel Density Estimation}
\newacronym{krr}{KRR}{Knowledge Representation and Reasoning}
\newacronym{owl}{OWL}{Web Ontology Language}
\newacronym{obo}{OBO}{Open Biological and Biomedical Ontology}
\newacronym{sw}{SW}{Semantic Web}
\newacronym{nmt}{NMT}{Neural Machine Translation}
\newacronym{kgc}{KGC}{Knowledge Graph Completion}
\newacronym{gcn}{GCN}{Graph Convolutional
Network}
\newacronym{qa}{QA}{Question Answering}
\newacronym{hitl}{HITL}{Human-in-the-loop}

\usepackage{nameref}
\usepackage{xspace}
\usepackage{csquotes}
\usepackage[capitalise]{cleveref}
\usepackage[final]{listings}
\usepackage{forest}
\forestset{%
my forest/.style={%
    for tree={%
            node options={text width=3cm, align=left},
            l sep=1cm, 
            parent anchor=south, 
            child anchor=north,
            if n children=0{tier=word}{},
        edge path={%
            \noexpand\path [\forestoption{edge}] (!u.parent anchor) -- +(0,-15pt) -| (.child anchor)\forestoption{edge label};
        },
    }
}
}

\usepackage{svg}
\usepackage{tikz}
\usepackage{pgfplots}
\pgfplotsset{compat=newest}
\usetikzlibrary{matrix,calc}
\usetikzlibrary{positioning}
\usepackage{algorithm}
\usepackage[noend]{algpseudocode}
\usepackage{diagbox}
\usepackage{tabularx,booktabs,ragged2e}
\newcolumntype{L}{>{\RaggedRight\arraybackslash}X}
\usepackage{menukeys} 

\usepackage{tabularx}
\lstdefinestyle{framed}{
  breaklines=true,
   basicstyle=\ttfamily,
  frame=single,
  alsoletter=:,
}

\usepackage{float}
\usepackage{nidanfloat}
\crefname{lstlisting}{listing}{listings}
\Crefname{lstlisting}{Listing}{Listings}

\usepackage{csquotes}

\usepackage{caption}
\usepackage{csquotes}
\usepackage{svg}
\usepackage{color}

\definecolor{lightgray}{rgb}{0.98,0.98,0.98}

\usepackage{amsmath}
 
\begin{document}

\maketitle

\begin{abstract}
Recent research efforts aiming to bridge the Neural-Symbolic gap for RDFS reasoning proved empirically that deep learning techniques can be used to learn RDFS inference rules. However, one of their main deficiencies compared to rule-based reasoners is the lack of derivations for the inferred triples (i.e. explainability in AI terms). In this paper, we build on these approaches to provide not only the inferred graph but also explain how these triples were inferred.  In the graph words approach, RDF graphs are represented as a sequence of graph words where inference can be achieved through neural machine translation. To achieve explainability in RDFS reasoning, we revisit this approach and introduce a new neural network model that gets the input graph--as a sequence of graph words-- as well as the encoding of the inferred triple and outputs the derivation for the inferred triple. We evaluated our justification model on two datasets: a synthetic dataset-- LUBM1 benchmark-- and a real-world dataset --ScholarlyData about conferences-- where the lowest validation accuracy approached 96\%. 
\end{abstract}

\section{Introduction}
The famous saying in the \gls{sw} community ``A Little Semantics Goes a Long Way''~\cite{littlesemantics} advocates for using light semantics such as \gls{rdfs}~\cite{w3crdfs} in order to achieve the \gls{sw} vision -- where humans and machines interact to process, produce and reason about the Web of data. \gls{rdfs} provides a way to describe the classes of \gls{rdf} resources, their hierarchy, the relations among the different resources and the hierarchy of these relations. Given an \gls{rdf} graph describing a set of resources (A-Box) and an \gls{rdfs} ontology describing the classes and relations (T-Box),  \gls{rdfs} entailment rules~\cite{w3crdfs} generate new facts that extend our knowledge about the A-Box.
\par In many applications such as \gls{qa}~\cite{DBLP:conf/www/UngerBLNGC12,DBLP:conf/sigmod/ZouHWYHZ14,DBLP:conf/clef/UngerFLNCCW14}, not only the inferred fact is required but also the derivation\footnote{Derivation and justification are used interchangeably in this paper.} of the triple. The derivation can be useful to justify the answer provided by the \gls{qa} system which allows the Human-in-the-loop to validate or reject the answer. An example of an \gls{rdfs} derivation in LUBM~\cite{lubm} is shown in~\cref{derivationExample}. 
\par The justification feature is within the evaluation criteria in \cite{Dentler2016ASO} where the authors survey ontology reasoners and list Pellet~\cite{pellet}, RACER~\cite{racer}, SWRL-IQ~\cite{swrliq} and CEL~\cite{cel} as ontology reasoners that provide justifications for the inferred triples. 
\begin{lstlisting}[label=derivationExample, escapeinside={(*}{*)}, caption={RDFS derivation example in LUBM}]
_:UndergraduateStudent46 rdf:type LUBM:Student .
  (*$\impliedby$*)
_:UndergraduateStudent46 	rdf:type 	LUBM:UndergraduateStudent .
LUBM:UndergraduateStudent	rdfs:subClassOf	LUBM:Student .
\end{lstlisting}

\par However, rule-based reasoners suffer from two main limitations:
\begin{enumerate}
    \item \textbf{Dealing with noisy data.} By design, rule-based reasoners are affected by noisy data and the veracity of the inferred triples is dependent on the veracity of the input triples. The Web being inherently noisy, even ``little semantics'' are susceptible to generate noisy inferences. In~\cite{dl4rdfs}, the authors draw a taxonomy of noise types in \gls{sw} data with respect to its effect on the inference.
    \item \textbf{Lack of support for approximate reasoning.} Rule-based reasoners aim at sound and complete inference. However, in many scenarios -- especially in interactive applications -- it is preferable to provide the user with an incomplete set of inferred triples in a timely manner rather than a ``delayed" complete inference. 
\end{enumerate} 
\par These two limitations motivated the exploration  of  deep learning techniques for \gls{sw} reasoning. In~\cite{hohenecker2017deep}, the authors propose \gls{rtn} as an adaptation for \gls{rntn}~\cite{DBLP:conf/nips/SocherCMN13} and report a speedup up to two orders of magnitude in the materialization. In~\cite{DBLP:journals/corr/abs-1811-04132}, the authors extend the End-to-end memory networks (MemN2N)~\cite{DBLP:conf/nips/SukhbaatarSWF15} to store \gls{rdf} triples and train the model to embed the knowledge graph and predict if a triple can be inferred from the input graph. \cite{dl4rdfs} focuses on the noise-tolerance aspect where the authors transform the \gls{rdf} graphs into a sequence of ``graph words" and model the \gls{rdfs} inference task as a machine translation task of graph words. 

\par Unfortunately, deep learning based reasoners suffer from a problem that is common for deep learning approaches in general: the lack of explainability. 
In~\cite{DBLP:journals/corr/abs-1708-08296}, the authors highlight four aspects for the need of ``Explainable AI", 1) verification of the system, 2) improvement of the system, 3) learning from the system and 4) compliance to legislation.
In the context of deep learning based reasoners, the lack of explainability translates into the inability to provide the derivation of the inferred triples. Towards combining the best of both worlds-- deep learning based reasoners' benefits (i.e.\ approximate reasoning and noise tolerance) as well as the justification feature in rule based reasoners-- this paper proposes a neural network model that learns the generation of justifications for the inferred triples.

\section{Approach Overview}
\label{sec:overview}

In order to input the RDF graph into the neural network-- that learns the generation of the inference graph and/or the justification of the inferred triples-- we first need to embed the RDF graph into a format that can be fed to neural networks. This phase is similar to the choice of the word embedding technique in deep learning for natural language understanding tasks. The literature contains many approaches for Knowledge graphs embedding  \cite{transe,wang2014knowledge,hole,conf/semweb/RistoskiP16,ji2015knowledge,trouillon2017knowledge,wang2017knowledge}. However only a few are tailored for \gls{rdfs} reasoning. One approach in particular is the ``graph words"~\cite{dl4rdfs} technique that converts an \gls{rdf} graph into a sequence of graph word IDs. Besides being tailored for \gls{rdfs} reasoning, the choice of this embedding technique is  motivated by the availability of the code and the expendability of the model as detailed further along.   
\par
\begin{figure*}[!ht]
\centering
\includegraphics[width=1.7\columnwidth]{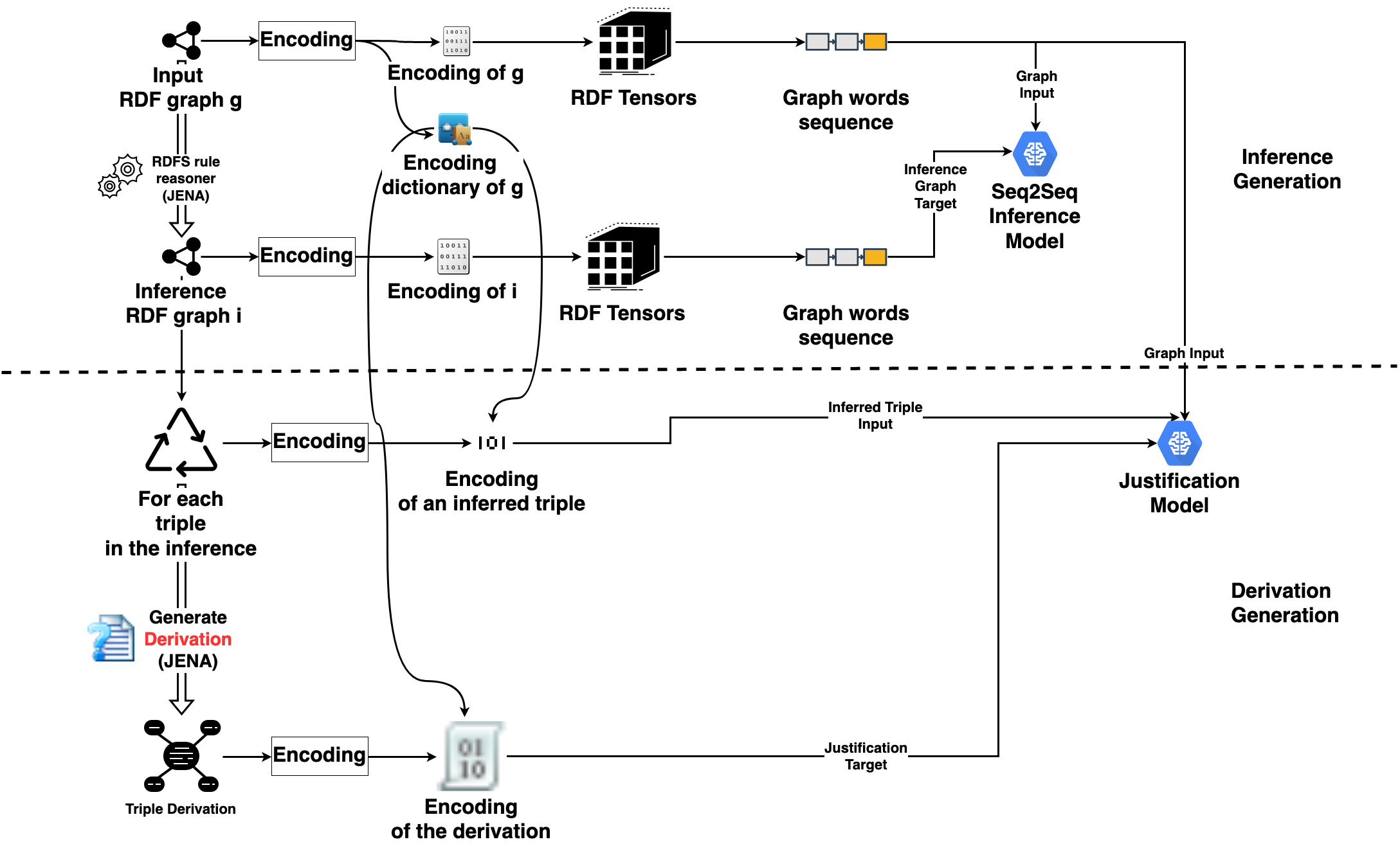}
\caption{Explainable Deep RDFS reasoner (Approach Overview)} 
\label{fig:sys_overview}
\end{figure*}

The augmentation of the graph words model to support the generation of the justifications for the inferred triples is illustrated in \cref{fig:sys_overview}. The upper segment depicts the graph words translation approach (Inference Generation) where each input graph is encoded into a 3D adjacency matrix (or Tensor). Each layer in the 3D adjacency matrix corresponds to the adjacency matrix between the \gls{rdf} resources which are linked according to a single predicate. These layers are then assigned an ID that represents their layout. The sequence of these IDs represents the \gls{rdf} graph hence the naming ``graph words". The training set becomes a parallel corpus between the sequence of graph words of the input graph and the sequence of graph words of the inference. The authors then trained a seq2seq~\cite{sutskever2014sequence} model that learns the inference generation.
\par The lower segment of~\cref{fig:sys_overview} (derivation generation) depicts the core contribution of this paper. Among the different explainability research efforts for deep learning networks, the closest to our approach is rationalization. Rationalization is defined in~\cite{harrison2017rationalization} as \begin{displayquote}[\cite{harrison2017rationalization}] a form of explanation that attempts to justify or explain an action or behavior based on how a human would explain a similar behavior. \end{displayquote} 
In practice, the rationalization of neural networks consists of training the neural network not only on the input and target data, but also on the rationale  behind the target data.
In our case, the justification is not provided by a human--rather it is provided by a rule-based reasoner (for instance JENA~\cite{jena}). The training input consists of the input \gls{rdf} graph represented as a sequence of graph words as well as the encoded inferred triple and the target is the encoded derivation. The justification model is a modified seq2seq model with two sequences as the input and one sequence as the output.

\section{RDF Data Representation}
\label{sec:prep}
In this section, we provide an overview of our data preparation process including ground-truthing and RDF tensors creation. 

\subsection{Ground-Truthing}
\begin{figure*}[!ht]
\centering
\includegraphics[width=1.5\columnwidth]{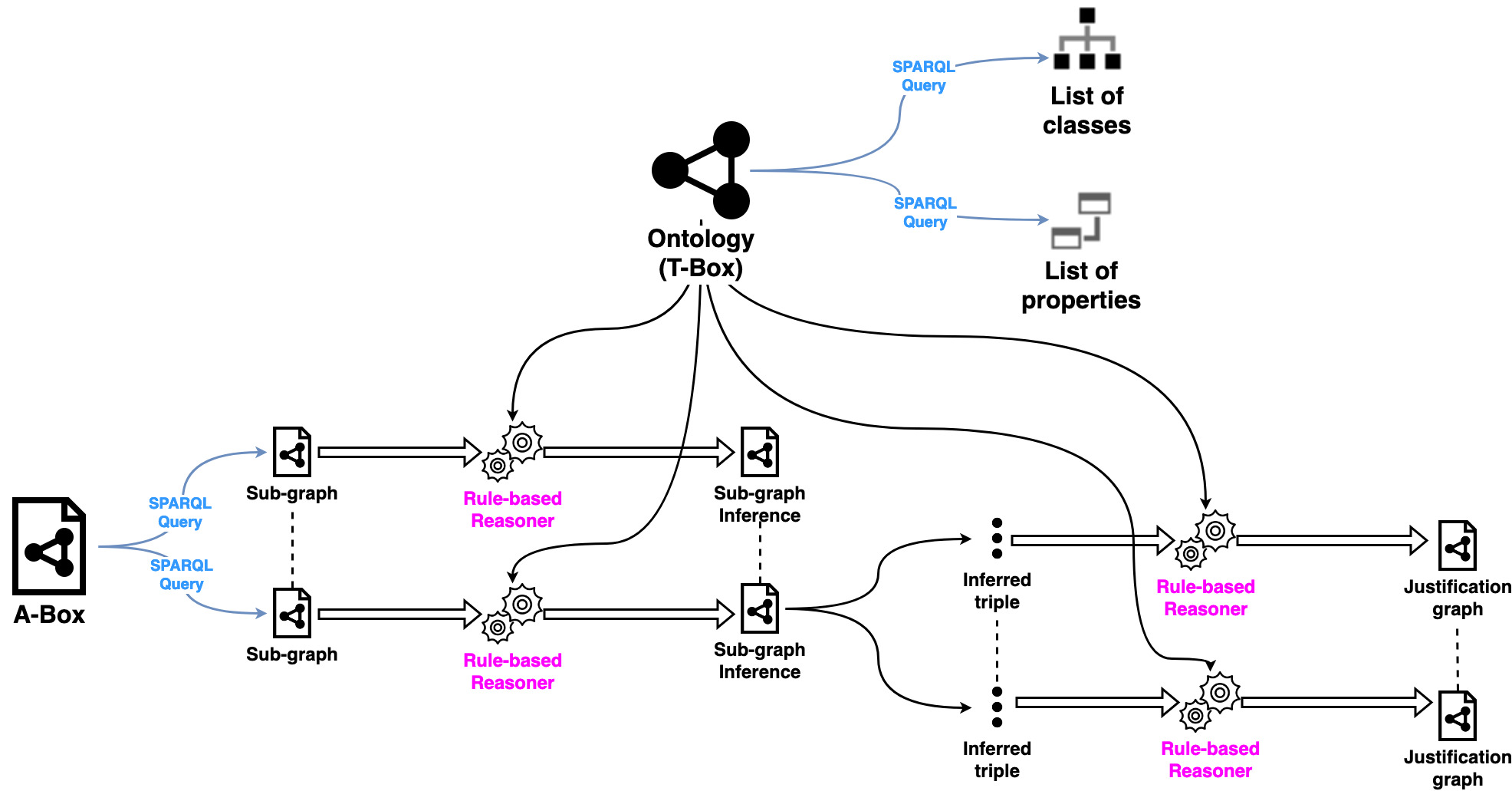}
\caption{Ground-Truthing pipeline} 
\label{fig:ground_truthing}
\end{figure*}

\cref{fig:ground_truthing} shows the overall pipeline of the ground-truthing process. This process is applied on the following datasets: \par
\stitle{\gls{lubm}}~\cite{lubm} is a well-known benchmark for evaluating RDF systems. Its ontology contains $42$ classes from the academic domain and $28$ properties describing these classes' relationships. In order to generate the ground truth, we split the LUBM1 dataset into $17,189$ sub-graphs where each graph represents the description (using SPARQL DESCRIBE) of a subject resource. 
The ground-truthing procedure consists of the following steps: 
\begin{enumerate}
\item Let $L$ be the set of these LUBM1 sub-graphs.
\item For each graph $l$ in $L$, we generate the inference $i$ of $l$ according to the LUBM ontology using JENA. 
\item For each inferred triple $t$ in the inference graph $i$, we generate the justification graph $j$ using JENA.
\item The training input is $(l,i)$ and the target is $j$
\end{enumerate}
This results into $78,766$ pairs of input graphs and inferred triples $(l,i)$. 
\par 
It is worth mentioning that the derivations provided by JENA are summarized derivations. This summarization is performed automatically by Jena and is not imposed by our pipeline. For instance, the justification of why an undergraduate student is of type \emph{Person} is simply because he is of type \emph{UnderGraduateStudent} and  \emph{UnderGraduateStudent} is a subClassOf \emph{Person} (\cref{derivationExample1}). However, in the LUBM ontology,    \emph{UnderGraduateStudent} is not a direct subClassof \emph{Person} but rather a subClassof \emph{Student} which is a subClassof \emph{Person} (\cref{derivationExample2}). This can be resultant of the ontology summarization~\cite{DBLP:conf/semweb/LiMd10} process.

\begin{lstlisting}[label=derivationExample1, escapeinside={(*}{*)}, caption={Summarized RDFS derivation}]
_:UndergraduateStudent46 rdf:type LUBM:Person .
  (*$\impliedby$*)
_:UndergraduateStudent46 	rdf:type 	LUBM:UndergraduateStudent .
LUBM:UndergraduateStudent	rdfs:subClassOf	LUBM:Person .
\end{lstlisting}

In the case of LUBM1, the summarized derivations generated by JENA contain either one or at most two triples, which will simplify the design of the justification neural network model. On the contrary, the justifications generated from the real-world datasets can be longer as detailed later.

\begin{lstlisting}[label=derivationExample2, escapeinside={(*}{*)}, caption={RDFS derivation}]
_:UndergraduateStudent46 rdf:type LUBM:Person .
  (*$\impliedby$*)
_:UndergraduateStudent46 	rdf:type 	LUBM:UndergraduateStudent .
LUBM:UndergraduateStudent	rdfs:subClassOf	LUBM:Student .
LUBM:Student	rdfs:subClassOf	LUBM:Person .
\end{lstlisting}

Finally, the generated data is split into $60\%$  training and $40\%$ validation sets. In this splitting phase, we used two stratification criteria--given that the input has two components--to ensure the balance of the training and validation sets. 
Otherwise there is a risk of having all the small classes in the training set, which will mistakenly inflate the accuracy. The first stratification criterion is based on the graph type (i.e.\ the type of the subject resource described in the graph). The second stratification criterion is based on the predicate and the object of the inferred triple. 

\stitle{ScholarlyData}~\cite{DBLP:conf/semweb/GentileN15} is a refactoring of the Semantic Web dog food database~\cite{DBLP:conf/semweb/MollerHHD07} that aims at keeping the database updated with the new conferences data and redesigning the T-Box according to the ontology design practices. The A-Box of the ScholarlyData contains $311,029$ triples that are split into $60,434$ sub-graphs using the same process as LUBM1. The main difference between both datasets resides in the length of the generated justifications. For instance, the justifications in ScholarlyData can contain more than two triples (see \cref{derivationExampleSC}). 

\begin{lstlisting}[label=derivationExampleSC, escapeinside={(*}{*)}, caption={Derivation example in ScholarlyData}]
_:iswc2010/paper/388 rdf:type SC:Document .
  (*$\impliedby$*)
_:iswc2010/paper/388 	SC:keyword 	"semantic web" .
_:iswc2010/paper/388 SC:abstract	"" .
_:iswc2010/paper/388 SC:title	"A Case Study of Linked Enterprise Data" .
\end{lstlisting}

\subsection{RDF Tensors Creation}
\label{sec:encoding}
The encoding and embedding stages aim to transform RDF graphs into a format suitable for neural network input. Given that the training data has three components to be encoded: the graph input, the inferred triple and the justification, we extended the encoding technique proposed in~\cite{dl4rdfs} (depicted in the upper segment of~\cref{fig:sys_overview}) to support the encoding of the derivations. The encoding technique in~\cite{dl4rdfs} can be briefly described in the following steps:
\begin{enumerate}
    \item Let $P$ be the set of ``active properties" which consists of the intersection between the properties in the ontology and the properties used in the A-Box. By using only the intersection and not the full set of properties in the ontology, the size of the embedding can be reduced dramatically without affecting the inference. 
    \item Let $P^{+}$ be the union of $P$ and \emph{rdf:type}.
    \item A fixed arbitrary order is imposed on the set $P^{+}$ throughout the training.
    \item Let $OP$ be the ordered list resulting from this sorting. Each property in $OP$ is assigned a integer ID $p_i$ according to its index.
    \item Let $GR$ be the set of global resources (i.e.\ classes in the ontology). An arbitrary order is also imposed on $GR$ and maintained throughout the training. Each class in $GR$ is assigned an integer ID $c_i$ according to its index. 
    \item Let $LR$ be the set of local resources (i.e.\ the \gls{rdf} resources that appear in the A-Box but not in the T-Box). Each resource in $LR$ is assigned an ID $l_i$ with an offset the size of $GR$. These IDs are reset when encoding the next \gls{rdf} graph.
\end{enumerate}
In this simplified encoding technique, the global $GR$ and local $LR$ resource sets are shared between the predicates. The authors also propose a more complex encoding technique where each group of predicates $p$ has its own $GR_p$ and $LR_p$ resource dictionaries.
\par By design, this encoding technique ensures that every resource that can appear in the inference has an ID assigned to it when encoding the input graph. This allowed us to use the same encoding technique for the input graph and the inferred triple. In contrast, the properties that can appear in the justification such as \emph{rdfs:subClassOf} and \emph{rdfs:subPropertyOf} are not present in $P^{+}$ as they are not used in the A-Box. As a remedy, we had to add the properties that are used in the \gls{rdfs} entailment rules~\cite{w3crdfs} to $P^{++}$ and similarly impose a fixed arbitrary order to assign an ID to each of these properties. Knowing that the derivation can contain two triples, an order is also imposed to sort these two triples according to the IDs of their predicates.
\par Imposing these lists of orders is mandatory for the training, otherwise the same inferred triple can have two targets as its justification depending on the order of the derivation triples. Keeping in mind that neural networks are functions' approximators~\cite{csaji2001approximation} but having the same input with two possible targets makes the problem outside function approximation realm.
\par At this stage, the input consists of two lists: 
\begin{description}
\item[Graph words sequence] A sequence of layers' IDs of size $P+$ where each layer ID represents the layout of the adjacency matrix according to the respective property.
\item[Inferred triple encoding] A sequence of 3 IDs for the subject, property and object of the inferred triple. These IDs are assigned according to the same encoding technique for the input graph.
\end{description}
The target consists of:
\begin{description}
\item[Derivation encoding] A padded sequence of size 6 corresponding to the IDs for each resource in the derivation.
\end{description}
Finally, the embedding of each layer in the graph words sequence is computed using \gls{hope}~\cite{hope}. For the inference and derivation sequences, we use one-hot vector encoding.

\section{Model Architecture}
\label{sec:model}
\begin{figure*}
    \centering
    \includegraphics[width=1.5\columnwidth]{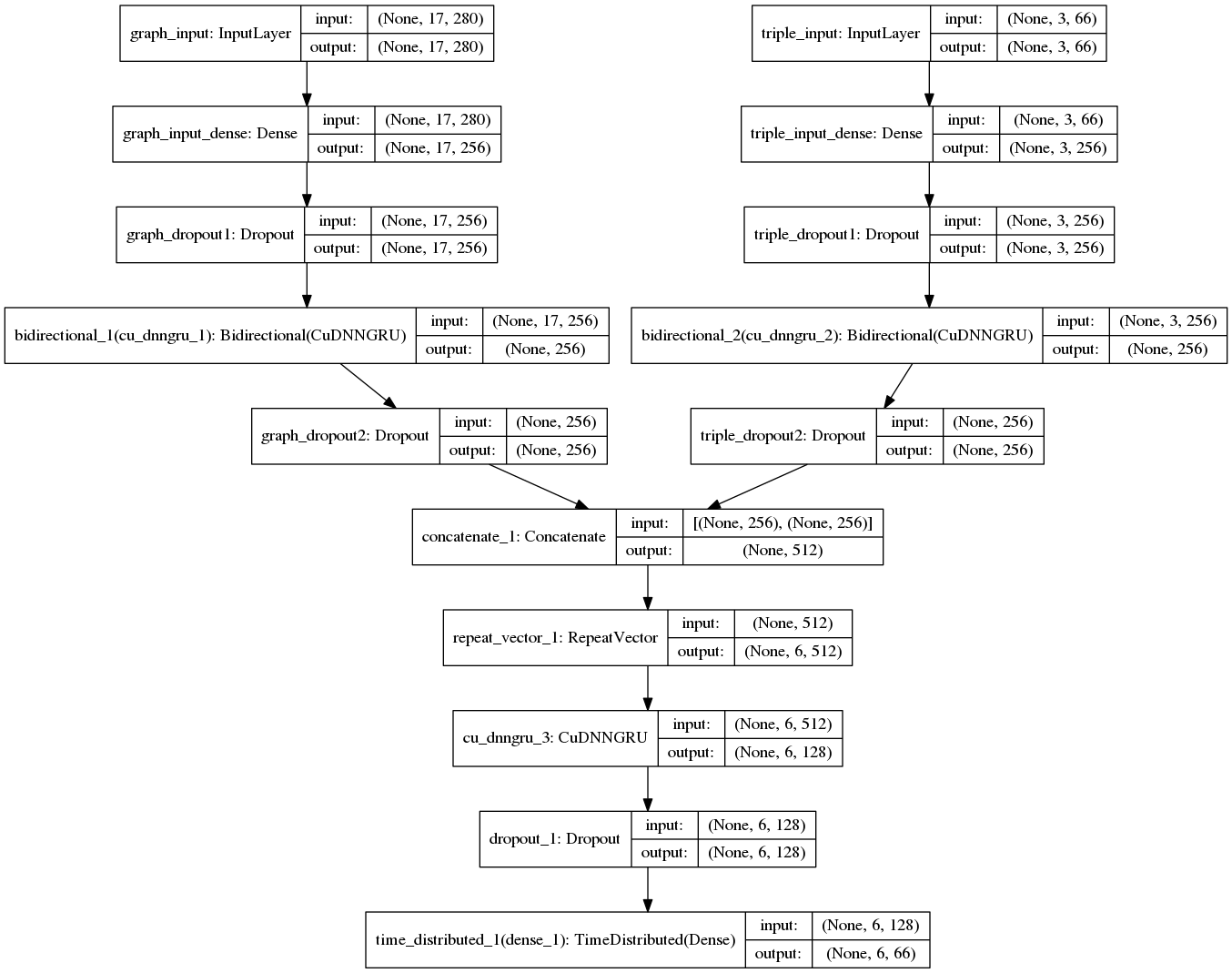}
    \caption{Model Architecture}
    \label{fig:fullmodel}
\end{figure*}

\cref{fig:fullmodel} shows the architecture of the justification neural network. It is a modified sequence-to-sequence model that takes two sequences as input. 
The graph input part is a tensor of size $(17,280)$ where $17$ represents the length of the sequence (i.e.\ the size of $P+$ in the case of LUBM). Each adjacency matrix in LUBM1 encoding is of size $70 \times 70$. When using the \gls{hope} embedding with a dimension of $4$, we reduce the size of the binary matrix to a real matrix of size $70 \times 4$ hence the $280$ parameter. \par 
The input triple tensor is of size $(3,66)$ where $66$ is the size of the one-hot vector. Resetting the local resources set $LR$ played a major role in reducing the size of this one-hot vector making the training faster and even possible. Each sequence (the graph sequence and the triple sequence) go then through similar sequence-to-sequence layers with a \gls{brnn}~\cite{schuster1997bidirectional} encoder which can be trained simultaneously in positive and negative time directions. The hidden representations are then merged in the \emph{concatenate} layer before being decoded into a sequence of $6$ items which should represent the derivation. Few \emph{dropout} layers are introduced in the architecture with a dropout factor of $0.2$ to prevent over-fitting and improve the generalization on the validation set.

\section{Evaluation}
\label{sec:eval}
This section shows the evaluation results for the task of providing justifications for the inferred triples in LUBM1 both using a baseline model and a full model architecture. Both models were trained and evaluated using two NVIDIA Tesla P100 GPUs used in parallel. 
LUBM1 dataset  has around one hundred thousand triples containing $17,189$ subject-resources within $15$ classes. The total number of pairs of graphs and inferred triples is $78,766$. It is worth noting that the same triple can be inferred by multiple graphs and the total number of inferred triples in LUBM1 is smaller.

\subsection{Baseline model}
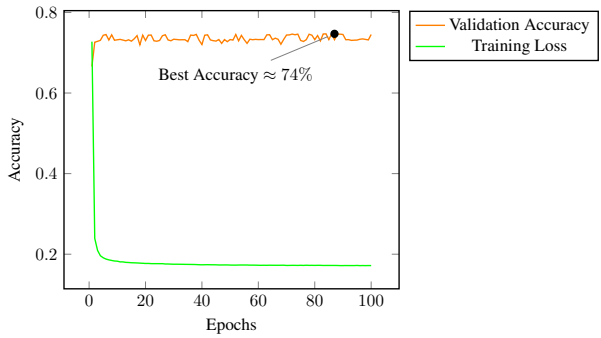
\begin{figure}
\centering
\begin{tikzpicture}[scale=0.65]
\begin{axis}[ xlabel=Epochs, ylabel=Accuracy, legend pos= outer north east, style=thick]  \addplot[color=orange] table[x=iteration,y=val_true_acc, col sep = comma] {explainablerdfsresults/baseline_training.txt};
\addplot[color=green] table[x=iteration,y=loss, col sep = comma] {explainablerdfsresults/baseline_training.txt};
\addplot[mark=*] coordinates {(87,0.746913384338643)} node[style=thick, pin=-120:{Best Accuracy $\approx 74\%$}]{} ;
\legend{Validation Accuracy,Training Loss,} \end{axis} 
\end{tikzpicture} 
\caption{Baseline model performance on LUBM1 dataset} 
\label{fig:base_model}
\end{figure}

Our first experiment shows the performance of a baseline model which gets as input only the inferred triple and outputs the justification. 
\cref{fig:base_model} shows training and validation accuracy of the baseline model. In the first few iterations, the training loss dropped significantly but stagnated afterwards. Similarly, the validation accuracy increased in the first few iterations 66\% to 73\% but improved very little later on. 

\par  It might be surprising that such a naive model can even reach this accuracy especially that the justification of an inferred triple depends on the input graph from which the inferred triple was derived. For example, if the question is why an entity is of type \emph{Person}, the answer can be: 1) because it is of type \emph{Student} \label{a} or 2) because it is of type \emph{Professor}.

Given that the majority of entities of type persons in LUBM1 are of type student and not professor, a naive model that always outputs justification \emph{Student} will be correct on the majority of the cases--thus the surprising relatively high accuracy and the stagnation when the model fails to generalize for smaller classes.

\subsection{Justification model}

\begin{figure}
\centering

\begin{tikzpicture}[scale=0.65] \begin{axis}[ xlabel=Epochs, ylabel=Accuracy, legend pos= outer north east, style=thick]  \addplot[color=orange] table[x=iteration,y=val_true_acc, col sep = comma] {explainablerdfsresults/full_model_training.txt};
\addplot[color=green] table[x=iteration,y=loss, col sep = comma] {explainablerdfsresults/full_model_training.txt};
\addplot[mark=*] coordinates {(88,0.959374107354468)} node[style=thick, pin=-120:{Best Accuracy $\approx 95.93\%$}]{} ;
\legend{Validation Accuracy,Training Loss,} 
\end{axis} 
\end{tikzpicture} 

\caption{Performance of our model on LUBM1 dataset} 
\label{fig:best_model}
\end{figure}
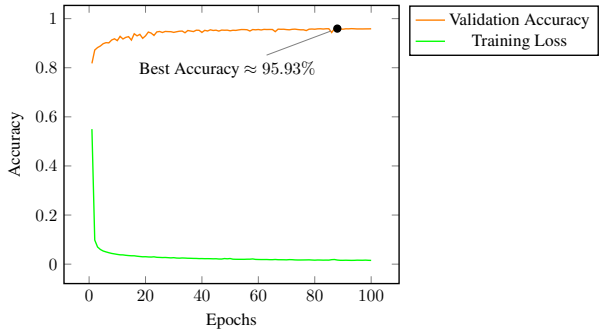
\begin{figure}[H]
\centering

\begin{tikzpicture}[scale=0.65] \begin{axis}[ xlabel=Epochs, ylabel=Accuracy, legend pos= outer north east, style=thick]  \addplot[color=orange] table[x=iteration,y=val_true_acc, col sep = comma] {explainablerdfsresults/full_model_training_scholarly.txt};
\addplot[color=green] table[x=iteration,y=loss, col sep = comma] {explainablerdfsresults/full_model_training_scholarly.txt};
\addplot[mark=*] coordinates {(88,0.979879261256065)} node[style=thick, pin=-120:{Best Accuracy $\approx 97.98\%$}]{} ;
\legend{Validation Accuracy,Training Loss,} 
\end{axis} 
\end{tikzpicture} 

\caption{Performance of our model on ScholarlyData} 
\label{fig:best_model2}
\end{figure}
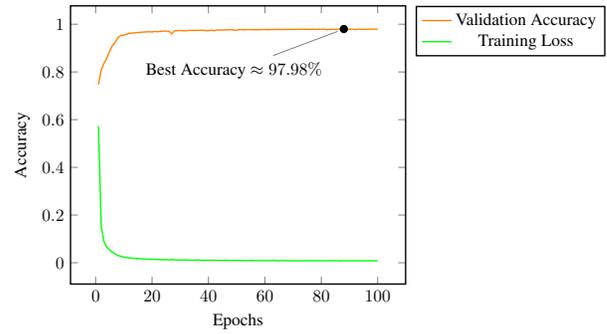

Our final model has a much better performance as shown in Figures~\ref{fig:best_model} and ~\ref{fig:best_model2} showing a validation accuracy of 95.93\% and 97.9\% for LUBM1 and Scholarly datasets respectively.

\subsection{Inference generation examples}
\Cref{derivationExamplegt} shows an example of ground truth derivation and \cref{derivationExamplep} shows an example where the justification model generates a wrong derivation by outputting the predicate \emph{undergraduateDegreeFrom} rather than \emph{doctoralDegreeFrom}. The model accuracy is computed with respect to the generated derivation by JENA. However there might be cases where the predicted derivation is different than JENA derivation but still valid. For instance if FullProfessor3 also got his undergraduate degree from the same university University879, the predicted derivation would have been valid.

\begin{lstlisting}[label=derivationExamplegt, escapeinside={(*}{*)}, caption={Ground truth derivation}]
_:FullProfessor3 rdf:type 	LUBM:Person .
  (*$\impliedby$*)
_:FullProfessor3 lubm:doctoralDegreeFrom 	<http://www.University879.edu> .
\end{lstlisting}

\begin{lstlisting}[label=derivationExamplep, escapeinside={(*}{*)}, caption={Predicted wrong derivation}]
_:FullProfessor3 rdf:type 	LUBM:Person .
  (*$\impliedby$*)
_:FullProfessor3 lubm:undergraduateDegreeFrom 	<http://www.University879.edu> .
\end{lstlisting}

\section{Related Work}\label{rw}

\stitle{Opening the black box of deep neural networks}
Deep learning models are often described as black box models as they are not built in a way that allows for the backtracking of their decision-making process. However, recent research efforts are attempting to demystify the hidden processing of deep learning networks and opening its black box such as ``Opening the Black Box of Deep Neural Networks via Information"~\cite{shwartz2017opening} and ``Rationalization: A Neural Machine Translation Approach to Generating Natural Language Explanations"~\cite{harrison2017rationalization}. In~\cite{harrison2017rationalization}, the authors train a deep model to play the video game Frogger. They also feed the  justifications generated by human players for each action. The model was able to not only make decisions that ultimately led to it winning the game, but it was also able to justify the sequence of actions it took.

\stitle{Deep Learning for Ontological Reasoning} 
\cite{dl4rdfs} showed that neural network translation can be utilized effectively in symbolic reasoning in the presence of noise. 
The authors proposed an embedding technique tailored for RDFS reasoning where graphs are represented as layers where each layout is encoded as a 3D adjacency matrix and forms a graph word. The input graph and its entailment are then represented as a sequence of graph words and the task of RDFS inference is then formulated as as translation task achieved through neural machine  translation.  In this paper, we build on this approach~\cite{dl4rdfs} to provide the explainability advantage for RDFS reasoning.  In~\cite{DBLP:journals/corr/abs-1811-04132}, the authors extend the End-to-end memory networks (MemN2N)~\cite{DBLP:conf/nips/SukhbaatarSWF15} to store \gls{rdf} triples and train the model to embed the knowledge graph and predict if a triple can be inferred from the input graph.

\section{Conclusion} 
\label{sec:conc}
In this paper, we aimed at combining the benefits of both deep learning reasoners as well as the justification feature in rule-based reasoners. We proposed a novel approach for achieving explanability in RDFS reasoning which is inspired by the rationalization technique for explainable AI. Both the RDF input graph as well as the inference and the derivation contribute to the input of the justification neural network. Our results show that this approach can provide a very high accuracy--up to 95.93\% on LUBM1. Possible extension of this  work is to test is on larger \gls{rdf} datasets and compare the derivation generation timing. As noise-tolerance was one of the motivations for proposing deep learning based reasoners, it will also be worth-while to test the justification generator model in the presence of noisy input data.

\bibliography{bibliography}
\bibliographystyle{aaai}
\balance

\end{document}